%% file: neurips_2026.tex
\documentclass{article}

\usepackage[final, main]{neurips_2026}

\usepackage[utf8]{inputenc} 
\usepackage[T1]{fontenc}    
\usepackage{hyperref}       
\usepackage{url}            
\usepackage{booktabs}       
\usepackage{amsfonts}       
\usepackage{nicefrac}       
\usepackage{microtype}      
\usepackage{xcolor}         
\usepackage{graphicx}
\usepackage{amsmath}
\usepackage{pifont}
\usepackage{algorithm}
\usepackage{algpseudocode}

\usepackage{multirow}
\usepackage{caption}
\usepackage{graphicx}
\usepackage{subcaption}

\usepackage{xcolor}
\usepackage{listings}
\usepackage[most]{tcolorbox}

\definecolor{llmgray}{HTML}{F7F7F7}
\definecolor{llmframe}{HTML}{D6D6D6}
\definecolor{llmblue}{HTML}{1F4E79}
\definecolor{llmtitletext}{HTML}{222222}
\definecolor{llmtitlebg}{HTML}{F0F0F0}

\lstdefinestyle{llmlog}{
  basicstyle=\ttfamily\scriptsize,
  breaklines=true,
  breakatwhitespace=false,
  columns=fullflexible,
  keepspaces=true,
  showstringspaces=false,
  upquote=true,
  xleftmargin=0pt,
  frame=none
}

\newtcblisting{llmlogbox}[2][]{%
  enhanced,
  breakable,
  colback=llmgray,
  colframe=llmframe,
  colbacktitle=llmtitlebg,
  coltitle=llmtitletext,
  boxrule=0.4pt,
  arc=2pt,
  left=1mm,
  right=1mm,
  top=1mm,
  bottom=1mm,
  fonttitle=\bfseries\small,
  coltitle=black,
  title={#2},
  listing only,
  listing options={style=llmlog},
  #1
}

\newcommand{\cmark}{\ding{51}}
\newcommand{\xmark}{\ding{55}}

\newcommand{\ourmethod}{\textsc{Cognitive Demand Steering}}
\newcommand{\ourmethodshort}{\textsc{CDS}}

\title{Cognitive Demand Steering for Adaptive Meta-Reasoning in Large Language Models}

\author{%
  John Scoville \\
  Thomson Reuters Foundational Research \\
  \And
  Shengzhuang Chen \\
  Thomson Reuters Foundational Research \\
  \And
  Yejin Bang \\
  Thomson Reuters Foundational Research \\
  \And
  Stefan Winzeck \\
  Thomson Reuters Foundational Research \\
  \And
  Jonathan Richard Schwarz \\
  Thomson Reuters Foundational Research \\
  Imperial College London} 
  
\begin{document}

\maketitle

\begin{abstract}
Recent meta-reasoning frameworks improve LLM reasoning by wrapping chain-of-thought generation in an iterative control loop, allowing more effective backtracking, termination of reasoning loops, and injection of promising reasoning patterns, among other strategy adjustments. Despite promising results, methods often rely on backward-looking reward functions, utilize coarse search actions, or require additional reasoning controller training requiring many-shot supervision. We introduce \ourmethod\   (\textsc{CDS}), a \textit{training-free} meta-reasoning framework equipped with residual demand assessment: at each step, an LLM-based progress evaluator characterizes the residual reasoning required to arrive at a solution rather than merely evaluating the previous step. This allows a meta-controller to select reasoning interventions comprising both general-purpose exemplars and actions (e.g., general guidance for quantitative reasoning) that directly tackle this forward-looking demand signal. This shift eliminates the need for any trained component while enabling zero-shot transfer across models and tasks with no adaptation. Rather than relying on coarse characterizations, we employ cognitive scales \citep{zhou2025generalscales} to both design interventions as well as profile initial problem complexity and residual demand signal over 16 dimensions motivated by cognitive science (e.g., attention and scan, learning and
abstraction, spatio-physical reasoning), giving the controller a fine-grained vocabulary for diagnosing. Averaged across three frontier LLMs and six reasoning benchmarks, \textsc{CDS} improves accuracy by $21.9\%$ over direct calls and $9\%$ over standard CoT reasoning, with the largest gains on difficult mathematics and coding tasks.

\end{abstract}

\input{sections/introduction}
\input{sections/related_work}

\input{sections/method}
\input{sections/experiments}

\input{sections/conclusion}
\input{sections/appendix}

\bibliographystyle{plain}
\bibliography{references}

\end{document}

%% file: sections/introduction.tex
\section{Introduction}

A recurring failure mode in Large Language Model (LLM) reasoning is a mismatch between the reasoning process and the demands of the task \cite{su2025between}. On challenging problems, a model may construct a plausible algebraic derivation while overlooking a decisive constraint, explore alternative solution paths before isolating the relevant information, or continue generating reasoning steps after the key uncertainty has already been resolved \cite{chen2024not, cuesta2025large}. These failure patterns suggest that the central challenge is not merely to elicit more reasoning from models, but to direct their reasoning toward the specific cognitive operations that remain unaddressed \cite{yan2025position}. As a concrete example, Figure~\ref{fig:arc191_d} shows a task (arc191-d from LiveCodeBench-Hard), where a textbook algorithm appears that it fits the task nicely, but upon examination of edge cases, is found to violate a key constraint of the problem.

\par
This observation motivates a line of work on adaptive inference, encompassing standard structured reasoning schemes such as Chain-of-Thought (CoT)~\cite{wei2022cot} (and models such as O1~\cite{openai2024reasoning} or R1~\cite{deepseek2025r1} trained for internal CoT via reinforcement learning), self-consistency~\cite{wang2022selfconsistency}, least-to-most prompting~\cite{zhou2022leasttomost}, and Tree-of-Thought (ToT)~\cite{yao2023tot}, as well as more explicitly controlled forms of meta-reasoning. 

Recent meta-reasoning systems select among reasoning strategies~\cite{gao2024meta}, decide whether to backtrack or restart~\cite{sui2025meta}, or transition between a discrete set of cognitive modes~\cite{jiang2026chain}. We refer to this family of approaches as \emph{action-centric} control: the controller is trained or prompted to map the current task context and reasoning state directly to a discrete intervention, such as selecting a reasoning method, switching cognitive mode, pruning a search branch, or restarting a reasoning trajectory. 

Despite the success of action-centric control in meta-reasoning, the approach still faces several limitations. We propose several improvements to improve the performance of meta-reasoning systems. 

First, \emph{compositional steering}. Complex tasks rarely demand a single monolithic reasoning style; they impose multiple interacting cognitive requirements, such as verbal comprehension, abstraction, logical deduction, quantitative verification, and domain knowledge retrieval. A controller that operates over a small discrete action set or a fixed set of modes may struggle both to represent such combinations explicitly and to elicit the corresponding mixture of reasoning behaviors from the model \cite{kargupta2025cognitive, song2026large}. In practice, this forces the controller towards selecting one coarse mode at a time, even when strong performance requires composing several distinct behaviors within the same reasoning trajectory. We propose to remedy this weakness by compositional steering, whereby the context passed to the model is injected with relevant behavioral modes to steer the reasoning process.\par

Second, \emph{demand-gap tracking}. A progress assessment is performed prior to reasoning, and once during each round. Progress towards goals is evaluated during each round of reasoning and residual demand is compared to the initial demand assessment. Low residual demand provides a signal to motivate early stopping. If the task is deemed solved, the progress assessment notes that the task is complete, terminating the reasoning loop to emit a final answer.\par

Third, \emph{adaptive compute allocation}. Without an explicit representation of unresolved task demands, decisions about whether to continue, verify, branch, or terminate are more likely to be driven by surface-level trajectory patterns, heuristic search policies, or the model's prior reasoning tendencies, rather than by a structured estimate of remaining task difficulty. We propose that compute can be allocated in an adaptive manner based on the residual demand assessment.\par

Fourth, \emph{adaptive verification targets}. Multi-step inference tasks risk propagating errors made in early stages, resulting in failure of the reasoning process. Without explicit checks to correct intermediate failures, significant downstream compute may be wasted in addition to producing a failed final outcome. We propose a dynamically generated list of adaptive verification targets (e.g. checking edge or corner cases, or inverting a result to verify its correctness) to introduce an error correcting mechanism to the meta-reasoning process.

In this paper, we propose \emph{Cognitive Demand Steering} (CDS), a training-free inference framework that reframes adaptive reasoning as \emph{demand tracking}, and directly addresses these limitations of action-centric control identified above. CDS can be regarded as a type of agentic loop using cognitive demand and assessed risk levels in order to inform unsupervised decisions on early stopping and iterative refinement. CDS first estimates a 16-dimensional cognitive demand profile for a given task, where each dimension is grounded in the cognitive scales taxonomy of~\cite{bean2025scalespp, zhou2025generalscales} and spans capabilities including verbal comprehension, relevant information identification, calibration of knowns and unknowns, logical reasoning, quantitative reasoning, and multi-domain knowledge retrieval. During inference, CDS iteratively evaluates the partial solution, estimates the degree to which each active dimension has been addressed, and computes the \emph{remaining cognitive demand}: the residual uncovered demand along each active dimension. This demand-gap signal serves as the controller's central state representation, in contrast to the discrete action sets or fixed mode switches of prior meta-reasoning systems. The demand gap drives effort and action planning in CDS, along with auxiliary risk levels assessed based on self-written validation tests, uncertainty estimation, and health checks on the reasoning process. By representing task demands and risks as a compositional, multi-dimensional profile rather than a discrete action, CDS enables the controller to simultaneously address multiple interacting cognitive requirements within a single trajectory, and to adaptively modulate both reasoning strategy and computational effort in proportion to the level of unresolved demand. Although this study is focused on training-free inference using CDS, the method is also suitable for generating reasoning trajectories for training models via supervised or reinforcement learning.\par

\begin{figure}[t]
  \centering
  \includegraphics[width=\textwidth]{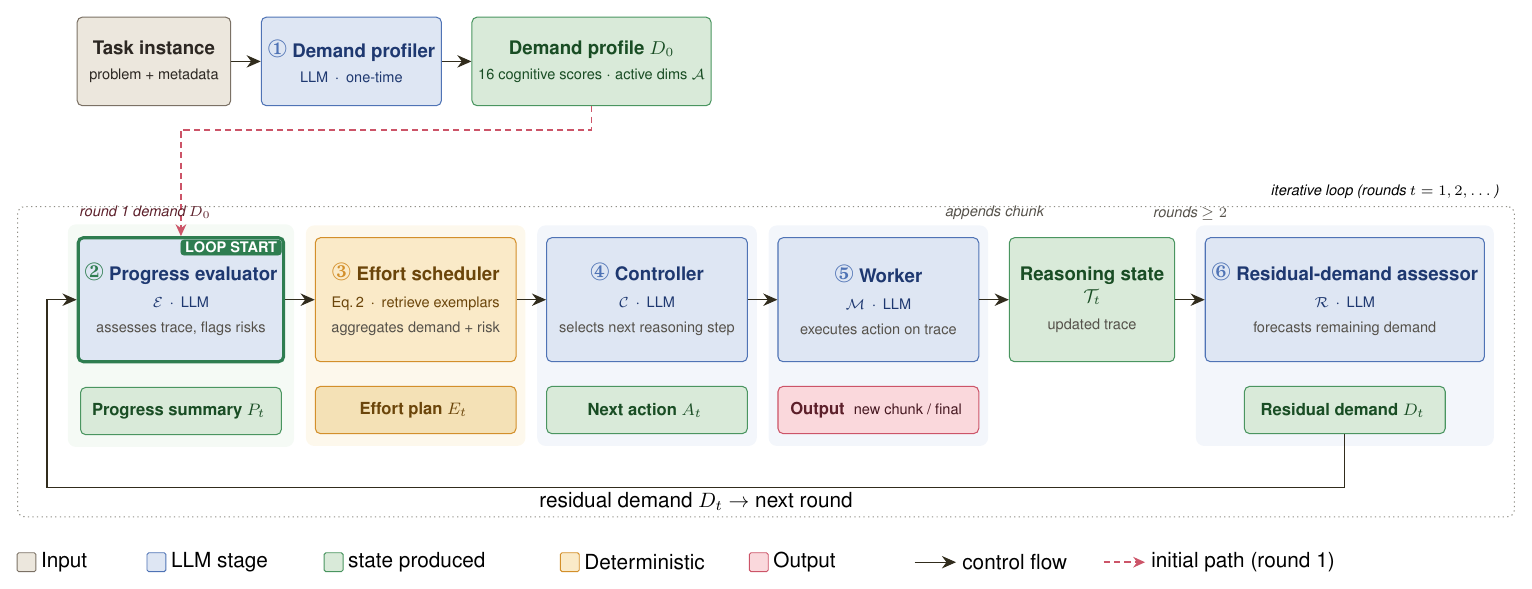}
  \caption{The CDS pipeline. A task is initially profiled by the demand profile to produce an initial demand profile and identify active cognitive scales dimensions. This demand profile is fed into the reasoning loop. Since no work has been done initially, no progress is reported, and residual demand reflects the full initial demand profile. Next, an effort level is planned based on the demand, and guidance is injected into context based on the active scales dimensions. Next, the controller chooses a next action based on the demand and current progress. The worker executes this plan and produces the next chunk of reasoning to append to the reasoning trace. The cycle continues until the progress evaluator deems the task complete, or if a maximum number of cycles have been reached.}
  \label{fig:figure1}
\end{figure}

We summarize our main contributions as follows:

\begin{itemize}
    \item \textbf{A demand-centered framework for adaptive reasoning.}
    We propose \textsc{Cognitive Demand Steering} (CDS), a training-free inference framework that replaces action-centric controller state with an explicit residual cognitive-demand state. CDS first profiles a task across cognitively grounded dimensions, then repeatedly estimates which demands remain unresolved. It tracks the overall cognitive demand, specific information or results that are missing, and any points needing explicit verification. It uses this signal to steer free-form controller actions, retrieve demand-specific guidance, and allocate reasoning effort.

    \item \textbf{Strong empirical performance across diverse reasoning tasks.}
    We evaluate CDS on six benchmarks covering mathematical reasoning, scientific question answering, and code generation, and across multiple frontier LLMs. CDS consistently improves over direct model calls and standard reasoning baselines, and is competitive with or stronger than prior meta-reasoning methods, with especially large gains on challenging tasks that require multi-step deduction, verification, or algorithmic planning.

    \item \textbf{Evidence that residual demand is a useful control signal.}
    Through iteration-scaling and ablation studies, we show that CDS benefits most from additional reasoning rounds when residual task demands remain high, while easier tasks saturate with fewer iterations. The demand gap helps to inform the progress evaluator when early stopping is advantageous, or when more effort for iterative refinement is needed. The ablations further indicate that demand-gap tracking provides an interpretable mechanism for diagnosing unresolved reasoning needs, though its effectiveness varies across backbone models.
\end{itemize}

\begin{table}[t]
  \caption{Comparison across three controller properties that are critical for an ideal adaptive reasoning framework: compositional steering, demand-gap tracking, and adaptive compute.}
  \label{tab:comparison}
  \centering
  \small
  \resizebox{\textwidth}{!}{%
  \begin{tabular}{lp{4.5cm}ccc}
    \toprule
    \textbf{Method} & \textbf{Control state} & \textbf{Training-free} & \textbf{Demand-gap tracking} & \textbf{Adaptive compute} \\
    \midrule
    MRP \cite{gao2024meta} & Task-level method choice & \cmark & \xmark & \xmark \\
    Meta-Reasoner \cite{sui2025meta} & Search/action state & \xmark & \xmark & \cmark \\
    CoM \cite{jiang2026chain} & Discrete mindset state & \cmark & \xmark & \cmark \\
    \midrule
    \textbf{CDS (Ours)} & \textbf{Residual demand state} & \textbf{\cmark} & \textbf{\cmark} & \textbf{\cmark} \\
    \bottomrule
  \end{tabular}
  }
\end{table}

%% file: sections/related_work.tex
\section{Related Work}

\textbf{Inference-time scaling}\quad
The first family of approaches improves reasoning by scaling test-time computation directly \cite{zhang2025survey}. CoT prompting \cite{wei2022cot} elicits intermediate reasoning traces; self-consistency \cite{wang2022selfconsistency} samples multiple traces and aggregates them; least-to-most prompting \cite{zhou2022leasttomost} decomposes harder tasks into progressively simpler subproblems; and ToT \cite{yao2023tot} introduces explicit branching and search over intermediate thoughts. Although these methods can improve reasoning quality, they fail to provide an explicit account of which reasoning remains unaddressed within a given trajectory. Since the efficacy of test-time compute scaling strategies is highly sensitive to the difficulty of the problem \cite{snell2025scaling}, their uniform, heuristic, or search-driven compute allocation is inherently suboptimal \cite{wu2025efficiency}. 

\par
Process reward models allow verification of intermediate results \cite{lightman2024let} rather than only the final results, allowing more granular analysis of the reasoning process, which in turn creates possibilities for corrective intervention. A second family of reasoning methods enables LLMs to improve their own outputs at inference time through iterative self-improvement. Self-Refine \cite{madaan2023selfrefine} generates feedback on completed drafts and revises them in a loop; Reflexion \cite{shinn2023reflexion} leverages verbal feedback across prior episodes to avoid repeating past mistakes.  Both approaches regulate inference, but they do so at a coarse granularity: Self-Refine operates on finished outputs, and Reflexion operates across separate trials rather than within a single trajectory. Critically, neither tracks the evolving structure of unresolved reasoning demands during solving. Indeed, frequent mid-trajectory reflection can actively disrupt reasoning flow, leading to underperformance on moderately complex tasks \cite{zhu2025scaling}. 

CDS shares the broader insight that inference should be regulated online, but differs in both the granularity and the object of regulation: rather than critiquing completed drafts or aggregating lessons across trials, we track demand gaps continuously within a single solving trajectory.

\par
\textbf{Adaptive meta-reasoners}\quad A growing body of work reframes reasoning around adaptivity-allocating reasoning effort to input characteristics such as difficulty and uncertainty \cite{wu2025efficiency}. The most closely related work treats reasoning itself as an object of control. Meta-Reasoning Prompting (MRP) \cite{gao2024meta} selects a reasoning method based on task characteristics. Meta-Reasoner \cite{sui2025meta} learns a policy over search actions such as backtracking, switching strategy, or restarting. The static implementation of Meta-Reasoner uses a fixed set of strategies, whereas the dynamic implementation of Meta-Reasoner can add new strategies to the search space during inference. MetaScale \cite{liu2025metascale} maintains a pool of \textit{meta-thoughts} - cognitive-strategy and problem-solving-mindset pairs - selecting among them with a multi-armed bandit and evolving high-reward ones via a genetic algorithm. Chain of Mindset (CoM) \cite{jiang2026chain} dynamically orchestrates a small set of distinct cognitive modes. Other methods aim to improve reasoning adaptively through consolidating cross-instance meta-cognitive experience from previous reasoning episodes \cite{zhuang2026beyond}, or dynamic selection of structural template to guide reasoning \cite{yang2406buffer}.

\par
CDS differs by maintaining an explicit, per-dimension estimate that tracks which aspects of the reasoning space have been addressed and which remain unresolved. CDS uses a reasoning loop similar to Meta-Reasoner but using an LLM controller rather than a trained multi-armed bandit, while explicitly tracking verification targets and missing information.

\par
\textbf{Cognitively grounded task representations}\quad More recently, tasks have been characterized through cognitively meaningful dimensions with the explicit purpose of enabling richer explanation and generalization beyond surface-level performance metrics. General Scales \cite{zhou2025generalscales} argues that placing task instances along non-saturating, continuous demand scales yields stronger explanatory and predictive power than black-box task representations, including for out-of-distribution prediction. Scales++ \cite{bean2025scalespp} extends this cognitive-scales perspective to the challenge of subset selection for efficient benchmark evaluation, prioritizing task demands over historical patterns of model failures. These works motivate an important premise of our method: cognitively grounded demand profiles can serve as interpretable, transferable representations of task structure, making them a plausible control state for adaptive reasoning. 

CDS repurposes this cognitive perspective to dynamically regulate reasoning at inference time. 
This lets the controller reason not only about \emph{what to do next}, but also about \emph{why more reasoning is needed}, \emph{which demand gaps remain open}, and \emph{which combination of reasoning behaviors should be synthesized next}.

\par

%% file: sections/method.tex
\section{Methodology}

\ourmethod~ operates a continuous feedback loop between a \textit{progress evaluator}, a \textit{residual demand assessor}, an \textit{effort scheduler}, a \textit{controller} and a \textit{worker} LLM. The core advancement is a training-free framework without the need for task or model adaptation. The core motivation is to have a forward-looking signal in the process loop in solving complex tasks, which enables dynamic guidance in reasoning.

\subsection{Architecture Overview}
\label{sec:overview}

\ourmethod~ operates as an iterative loop around an unmodified worker LLM $\mathcal{M}$ that is prompted to show its reasoning explicitly rather than relying on hidden chain-of-thought. Given a task instance $q$, the framework first runs a one-time \textbf{demand profiler} (\S\ref{subsec:demand}) that scores the instance across 16 dimensions motivated by cognitive sciences \cite{zhou2025generalscales} and selects the top-$k$ highest-demand dimensions as the active working set. Then, the framework proceeds in rounds $t = 1, 2, \dots$ Each round comprises five stages:
\begin{enumerate}
    \item \textbf{Progress evaluator} $\mathcal{E}$  (\S\ref{subsec:progress}) summarises the current reasoning state and flags progress completion, reasoning risks and uncertainty level.
    \item \textbf{Residual demand assessor} $\mathcal{R}$  (\S\ref{subsec:residual}) estimates what demands the problem \emph{still} places on the solver. And, it provides guidance in the form of general purpose exemplars\footnote{Example: critical thinking - Audit the reasoning before advancing. Look for unsupported assumptions, fragile steps, off-by-one errors, unit mismatches, invalid simplifications, and whether an alternative interpretation could break the answer.} (listed in Appendix \ref{subsec:exemplars}) providing advice and warnings about the active working set of dimensions.
    \item \textbf{Effort scheduler} (\S\ref{subsec:effort}) Converts the residual demand assessment into computational effort levels for the current task.
    \item \textbf{Controller} $\mathcal{C}$ (\S\ref{subsec:controller}) ingests the residual demand profile together with retrieved guidance and selects the next reasoning intervention in free-form natural language.
    \item \textbf{Worker} $\mathcal{M}$ (\S\ref{subsec:loop}) executes the intervention, producing the next segment of the reasoning trace or, if it judges the solved problem, a final answer.
\end{enumerate}
The output of the worker is appended to a running reasoning trace $\mathcal{T}_t$ and fed back into the progress evaluator, closing the loop.
Figure~\ref{fig:figure1} illustrates the full pipeline.

\ourmethod~has no reward function and no task-specific training. Instead, the residual demand assessment provides a \emph{forward-looking} signal that an LLM controller can act on zero-shot. Because both assessment and control are performed in natural language by a general-purpose LLM, the framework is transferred across tasks and backbone models without adaptation.

\subsection{Pseudocode}

Pseudocode for \ourmethod\ can be seen in Algorithm \ref{alg:cds}, and pseudocode for its subroutines may be found in Algorithm \ref{alg:subroutines}.

\begin{algorithm}[t]
\caption{Cognitive Demand Steering (\textsc{Cds})}
\label{alg:cds}
\begin{algorithmic}[1]
\Require task $x$; LLM backend $M$; demand taxonomy $\mathcal{D}$ ($16$ cognitive/knowledge dims);
         max rounds $R$; effort-tiered worker backends $\{M_\tau\}_{\tau=1}^{5}$
\Ensure  reasoning trace $\rho$ ending in an extractable \texttt{FINAL ANSWER}
\Statex
\State $\pi \gets \textsc{ProfileTask}(x)$
   \Comment{General scales: demand $s\!:\!\mathcal{D}\!\to\!\{0,\dots,5\}$, top dims, initial tier $\tau_0$}
\State $r \gets \pi.\textsc{Demand}(d)\ \forall d \in \mathcal{A}$
   \Comment{bootstrap residual demand from initial profile}
\State $\mathcal{A} \gets \textsc{TopK}(r,\,4)$
   \Comment{active demand dimensions to steer toward}
\State $\rho \gets \varepsilon$;\quad $\textsc{trace} \gets \emptyset$
   \Comment{$\varepsilon$ = empty reasoning}
\For{$t \gets 0$ \textbf{to} $R-1$}
    \State $p \gets \textsc{EvaluateProgress}(x,\rho,\pi,\mathcal{A})$
    \If{$p.\textsc{solved}$}
        \State \Return $\textsc{Finalize}(x,\rho,p)$ \Comment{early stop, pre-worker}
    \EndIf
    \State $e \gets \textsc{ScheduleEffort}(\pi,p,r)$
        \Comment{tier $\tau$ + step / branch / verification budget}
    \State $\mathcal{A}^\star \gets \textsc{TopK}(r,\,3)$ \Comment{subset of highest residual-demand dims}
    \State $E \gets \textsc{RetrieveExemplars}(\mathcal{A}^\star,\,r)$
        \Comment{Retrieve guidance for dims with $r_d > \theta$}
    \State $g \gets \textsc{Controller}(x,\mathcal{A}^\star,r,p,e,E)$
        \Comment{steering signal: next action, gaps, checks, style}
    \State $M_\tau \gets \textsc{SelectWorker}(e.\textsc{tier})$
    \State $w \gets \textsc{Worker}(M_\tau,\,x,\,\rho,\,g,\,e.\textsc{tier})$
        \Comment{one demand-steered reasoning chunk}
    \State $\rho \gets \rho \,\Vert\, w$
    \State $r \gets \textsc{ResidualDemand}(x,\pi,p,\mathcal{A})$
        \Comment{remaining demand $r_d\!\in\![0,5]\ \forall d\!\in\!\mathcal{A}$}
\EndFor
\State \Return $\rho$
\end{algorithmic}
\end{algorithm}

\begin{algorithm}[t]
\caption{\textsc{Cds} subroutines (LLM oracle queries on $M$)}
\label{alg:subroutines}
\begin{algorithmic}[1]
\Function{ResidualDemand}{$x,\pi,p,\mathcal{A}$}
   
    \State $\{r_d\} \gets M\bigl(\textsc{ResidualDemandPrompt}(x,\mathcal{A},s,p)\bigr)$
    \State \textbf{return} $\bigl\{\,\mathrm{clip}\!\bigl(r_d \;\textbf{else}\; \hat r_d,\,0,\,5\bigr)\,\bigr\}_{d\in\mathcal{A}}$
\EndFunction
\Statex
\Function{ScheduleEffort}{$\pi,p,r$}
    \State $T \gets \textsc{TopK}(r, 3)$
    \State $u \gets 5 (p.\mathrm{uncertainty})$
    \State $c \gets 5 \max(p.\mathrm{loop}, p.\mathrm{contradiction})$
    \State $\tau \gets \operatorname{round}\!\left(
        0.45\,\max(T) + 0.25\,\overline{T}
        + 0.15\,u + 0.15\,c
    \right)$
\EndFunction
\Statex
\Function{EvaluateProgress}{$x,\rho,\pi,\mathcal{A}$}
    \State $\textsc{ProgressState} \gets M\bigl(\textsc{ProgressPrompt}(x,\rho,\mathcal{A},s)\bigr)$
    \State \textbf{return} $\textsc{ProgressState}$
\EndFunction
\Statex
\Function{Controller}{$x,\mathcal{A}^\star,r,p,e,E$}
    \State $g \gets M\bigl(\textsc{ControllerPrompt}(x,\mathcal{A}^\star,r,p,e.\mathrm{tier},E)\bigr)$
    \State $g.\{\textsc{activeDims},\textsc{exemplars}\} \gets \{\mathcal{A}^\star, E\}$
    \State \textbf{return} $g$
\EndFunction
\Statex
\Function{Worker}{$M_\tau,x,\rho,g,\tau$}
    \State \textbf{return} $M_\tau\bigl(\textsc{WorkerPrompt}(x,\rho,g,\tau)\bigr)$
        \Comment{next chunk only; obeys steering $g$ \& budget $\tau$}
\EndFunction
\end{algorithmic}
\end{algorithm}

\subsection{Initial Cognitive Demand Profiling}
\label{subsec:demand}

Before the iterative loop begins, \ourmethod~constructs an initial \emph{demand profile} for the task instance $q$.
A demand profiler prompt presents $q$ to an LLM and elicits scores across $K{=}16$ cognitive and knowledge dimensions, each rated on an integer scale from 0 (no demand) to 5 (extreme demand). The dimensions span capabilities that recur across reasoning-intensive tasks--- attention and scan, calibrating knowns and unknowns, abstraction, critical thinking, identifying relevant information, logical reasoning, quantitative reasoning, verbal comprehension, and multiple knowledge domains (the full set is listed in Appendix~\ref{app:dimensions}).

The profile serves two purposes: it gives the residual demand assessment a fixed set of axes along which to track progress, and it enables selective retrieval of dimension-specific guidance (\S\ref{subsec:exemplars}).
Not all 16 dimensions are relevant to every problem, so we retain only the top-$k$ highest-scoring dimensions as the \emph{active working set} $\mathcal{A} \subset \{1, \dots, K\}$, with $k{=}4$ fixed throughout our experiments, this value was obtained via a hyperparameter search. Moreover, the active-set dimension with the lowest demand at each step is omitted from the working set.
 
\subsection{Progress Evaluator}
\label{subsec:progress}

\[
  \mathcal{E}(q,\ \mathcal{T}_t,\ \mathcal{A})
  \;\longrightarrow\;
  P_t = \bigl\{\, c_t,\ l_t,\ u_t,\ v_t,\ m_t,\ \textit{task\_solved} \,\bigr\}
\]

At each round $t$, the progress evaluator $\mathcal{E}$ receives the task instance $q$, the current reasoning trace $\mathcal{T}_t$ and active dimensions $\mathcal{A}$. It produces the progress summary $P_t$, including task completion, reasoning risks and uncertainty level. On the first iteration ($t{=}1$), the trace is empty and $P_1$ is initialised to a null state: uncertainty $u_1 = 0.5$, and all risk scores set to zero.

The progress assessor also compiles a list of useful \textbf{verification targets} $v_t$ (e.g. edge cases, unit tests) and \textbf{missing information} (e.g. missing facts or needed intermediate results) $m_t$ during each reasoning round, these are incorporated into the context and passed to the controller. Verification targets are descriptions of natural language tests on the reasoning process, they are generated by the progress assessor and used by the controller to plan validation checks on the reasoning process. These checks are subseqently executed by the worker. The effort budget determines the level of verification effort planned by the controller, and the results of checking verification targets play a key role in assessing risk levels. Missing information is also identified by the progress assessor and informs the controller's planning regarding information that needs to be obtained in future reasoning rounds. 

For $t > 1$, the evaluator diagnoses three signals that capture the dominant failure modes, motivated by \cite{sui2025meta}: (i) \textbf{Contradiction risk} $c_t \in [0, 1]$: whether the current trace contains internal inconsistencies (e.g. from a missed verification target) (ii) \textbf{Loop risk} $l_t \in [0, 1]$: whether the reasoning has entered a repetitive or circular pattern (iii)\textbf{Uncertainty} $u_t \in [0, 1]$. 
Here, uncertainty is the progress assessor's assessment of confidence or uncertainty in the reasoning process. Contradiction risk and loop risk are passed to the effort scheduler (\S\ref{subsec:effort}).
Together, $u_t$ and $r_t$ provide the progress evaluator's contribution to the effort computation (Eq.~\ref{eq:effort}): high uncertainty or instability drives effort upward, triggering more cautious reasoning in subsequent rounds.

 \subsection{Residual Demand Assessor}
\label{subsec:residual}

Given the progress  $P_t$ and the initial demand profile $D_0 = \{d_0^{(i)}\}_{i \in \mathcal{A}}$ restricted to the active dimensions, the residual demand assessor produces a step-specific demand profile:

\begin{equation}
    D_t = \mathcal{R}(q, P_t, D_0),
    \label{eq:residual}
\end{equation}
where $D_t = \{d_t^{(i)}\}_{i \in \mathcal{A}}$ records the remaining demand along each active dimension at round $t$.

As reasoning progresses, dimensions whose demands have been satisfied see their scores decrease toward zero. The residual demand assessment is predictive as it analyses what the problem still needs in order to be solved. Then, the controller can make the decisions on what actions/strategies need to be done based on the needs of the task.

For each active dimension $i \in \mathcal{A}$, 
\ourmethod\ injects a small set of general exemplars into the controller's prompt context. These exemplars are pre-authored in a static dictionary $\mathcal{G} = \{i \mapsto \{g_{i}\}\}$.

\subsection{Effort Scheduling}
\label{subsec:effort}
 
The residual demand profile $D_t$ characterises \emph{what} the problem still needs; the effort scheduler translates this into \emph{how hard} the next reasoning step should work.
We compute a scalar effort score $E_t \in [0, 5]$ that aggregates the residual demand with risk signals from the progress evaluator:

\begin{equation}
    E_t = \alpha_0 \cdot \max_{i \in \mathcal{A}} d_t^{(i)} \;+\; \alpha_1 \cdot \overline{D}_t \;+\; \alpha_2 \cdot 5 \cdot u_t \;+\; \alpha_3 \cdot 5 \cdot r_t,
    \label{eq:effort}
\end{equation}
where $\overline{D}_t = \frac{1}{|\mathcal{A}|}\sum_{i \in \mathcal{A}} d_t^{(i)}$ is the mean residual demand over active dimensions, $u_t$ is the uncertainty estimate from the progress evaluator, $r_t$ is the instability score derived from the progress evaluator's contradiction and loop risk scores, and the $\alpha_i$ are weights of each of these terms. $\alpha=(0.45, 0.25, 0.15, 0.15)$ in our evaluations. $r_t$ and $u_t$ are multiplied by 5 so all scores are in the range $[0, 5]$.
The weighting assigns the largest influence to the hardest remaining dimension, ensuring that a single unresolved bottleneck keeps effort high even if other dimensions are nearly satisfied.

The continuous score is mapped to a discrete effort level $\ell_t \in \{1, \dots, 5\}$ that determines the reasoning guidance injected into the controller and worker prompts. $\text{effortlevel}_t = \max\!\left(1,\ \min\!\left(5,\ \lfloor E_t \rceil\right)\right)$.

 %
 
Each level prescribes increasingly thorough reasoning behaviour: lower levels permit the worker to proceed with minimal overhead, while higher levels instruct the controller to request verification steps, decomposition, or alternative solution paths, and direct the worker toward more cautious execution. This creates a natural form of \emph{adaptive compute allocation}--- easy subproblems or nearly-solved instances consume fewer tokens and iterations, while hard or error-prone states trigger deeper reasoning automatically. It is possible to use the effort schedule to route higher-effort steps to larger or stronger models, but our evaluations use a single model per test with an adaptive compute budget. 

\subsection{Controller}
\label{subsec:controller}

The controller $\mathcal{C}$ is a decision maker in the framework, given all gathered context from previous steps. It produces a natural-language \emph{action} $A_t$ that directs the worker's next step.

Importantly, the action is free-form -- there is no constraint for the controller to select among a fixed set of actions. As such, the controller is unconstrained in its choice of next actions, and it is free to choose a next action that is detailed in content and specific to the current state of the task. For example, a next action for a math problem could be ``Substitute 1/z with the conjugate of z divided by 16, simplify the expression, extract the real part using trigonometric forms, and find its maximum amplitude.'' In contrast to the small, discrete set of Meta-Reasoner (\cite{sui2025meta}) strategies, this open-ended action space enables the full space of possible action guidance based on what problem demands. It is implemented with an LLM that does not require any parameter updates: it solely conditions on task, progress summary, schedule effort, residual demand profile and on the exemplars.

\subsection{Worker and Loop Termination}
\label{subsec:loop}

The worker $\mathcal{M}$ intakes the action $A_t$ and Effort $E_t$ from the controller along with the accumulated context and full reasoning trace $\mathcal{T}_{t-1}$ and produces the next reasoning segment $s_t$. If a maximum count of reasoning iterations has been exceeded, the loop terminates the reasoning trace. Alternatively, the loop terminates the trace once the problem is judged as solved by the progress assessor. Otherwise, the new reasoning chunk appended to the trace $T$ and iteration continues.

%% file: sections/experiments.tex
\section{Experiments}
We evaluate \ourmethod\ on a suite of reasoning benchmarks designed to stress different combinations of cognitive demands, including mathematical deduction, algorithmic execution, scientific knowledge integration, and code synthesis. Additionally, we study how the predictive power of the reasoning trace increases with the number of reasoning rounds, and consider the token efficiency of the different methods. Our evaluation asks three main questions:

\begin{itemize}
\item Does \ourmethodshort\ improve reasoning accuracy over direct inference and standard inference-time reasoning methods?
\item Does explicit residual-demand tracking provide benefits beyond generic iterative reasoning or adaptive action selection?
\item How does performance vary as the number of reasoning rounds increases, and does \ourmethodshort\ allocate additional reasoning effort most effectively on harder tasks?
\end{itemize}

\begin{table*}[t]
\centering
\caption{Performance comparison of reasoning methods across models and benchmarks. Results are reported as percentage (\%) scores. LCB = LiveCodeBench (Easy/Medium/Hard). Meta-Reasoner-S = Meta Reasoner Static, Meta-Reasoner-D = Meta Reasoner Dynamic. Best and 2nd best result per benchmark per model marked \textbf{bold} or underlined, respectively.} 
\label{tab:main_results}
\resizebox{0.9\textwidth}{!}{%
\begin{tabular}{lcccccc|c}
\toprule
\textbf{Method} & \textbf{AIME} & \textbf{MATH500} & \textbf{GPQA} & \textbf{LCB-Easy} & \textbf{LCB-Med} & \textbf{LCB-Hard} & \textbf{Average} \\
\midrule
\multicolumn{8}{l}{\textit{Gemini 3.1 Pro}} \\
\midrule
Direct          & 55.00 & 90.80 & 88.38 & 78.88 & 54.05 & 32.86 & 66.66 \\
CoT             & 65.00 & 91.40 & 88.89 & 78.88 & 55.35 & 39.71 & 69.87 \\
ToT             & 90.00 & 91.20 & 91.41 & \textbf{84.78} & \textbf{65.80} & 63.14 & 81.06 \\
Meta-Reasoner-S & \underline{95.00} & \textbf{93.00} & 92.42 & 77.64 & 62.92 & 63.14 & 80.69 \\
Meta-Reasoner-D & \underline{95.00} & \underline{92.60} & \underline{94.44} & 76.40 & 63.45 & \underline{69.14} & \underline{81.84}\\
\ourmethodshort\ (Ours)      & \textbf{98.33} & 91.40 & \textbf{95.45} & \underline{82.30} &    \underline{65.01} &  \textbf{72.57} &  \textbf{84.18} \\
\midrule
\multicolumn{8}{l}{\textit{Claude Sonnet 4.6}} \\
\midrule
Direct          & 35.00 & 80.80 & 72.22 & 83.23 & 46.48 & 17.14 & 55.81 \\
CoT             & 73.33 & 87.60 & 72.22 & 78.26 & 62.92 & 45.71 & 70.01 \\
ToT             & 66.67 & 78.00 & 78.79 & \textbf{84.47} & \textbf{65.27} & \textbf{66.86} & 73.34 \\
Meta-Reasoner-S & 76.67 & \textbf{88.80} & \textbf{83.33} & 83.54 & \underline{63.97} & 56.57 & 75.48 \\
Meta-Reasoner-D & \underline{86.67} & \underline{88.60} & 79.29 & \textbf{84.47 }& 63.45 & 60.29 &\underline{77.13} \\
\ourmethodshort\ (Ours)      & \textbf{96.67} & 87.60 & \underline{81.82} & \underline{84.16 }&    63.71 &  \underline{61.43} & \textbf{79.23}\\
\midrule
\multicolumn{8}{l}{\textit{GPT-5.4}} \\
\midrule
Direct          & 5.00 & 41.20 & 50.00 & 83.85 & 62.92 & 50.29 & 48.88 \\
CoT             & \underline{61.67} & \underline{82.40} & 72.22 & 81.06 & 61.88 & 50.00 & 68.20 \\
ToT             & 20.00 & 53.20 & 69.70 & \textbf{84.16} & \textbf{64.75} &\textbf{ 62.57} & 59.06 \\
Meta-Reasoner-S & 51.67 & \textbf{84.00} &\underline{ 74.24 }& \underline{83.85} &\underline{ 63.45 }& 50.86 & 68.01 \\
Meta-Reasoner-D & 56.67 & \textbf{84.00} & 72.73 & \underline{83.85} & 61.88 & 57.43 & \underline{69.43} \\
\ourmethodshort\ (Ours)      & \textbf{70.00} &    82.00 &    \textbf{75.25} &    81.68 &    62.66 & \underline{59.43} & \textbf{71.84}\\
\bottomrule
\end{tabular}%
}
\end{table*}

\subsection{Experiment Setup}
\paragraph{Datasets}
We evaluate \ourmethodshort~on a diverse suite of challenging reasoning benchmarks and report accuracies:
\begin{itemize}
    \item \textbf{Mathematics:} MATH500 \cite{lightman2024let} and AIME combined 2024/2025 mathematics exercises to test deep logical deduction and algorithmic execution.
    \item \textbf{Coding:} LiveCodeBench v6 code generation tasks \cite{livecodebench} to evaluate multi-step algorithmic reasoning and code generation, split into Easy/Medium/Hard categories to demonstrate performance on coding tasks that vary in difficulty. 
    \item \textbf{Scientific \& General QA:} GPQA Diamond \cite{gpqa} problems to assess domain knowledge integration and knowledge-intensive reasoning performance.
\end{itemize}

We allow up to 12 iterations of the \ourmethodshort~loop in the evaluations. No tool use is allowed. LiveCodeBench Code is evaluated on an Amazon Sagemaker ml.t3.xlarge instance for 3 seconds to judge correctness.

\paragraph{Baselines}
We consider various prompting methods as baseline, including direct model call (zero-shot; Direct), Chain-of-Thought (CoT)~\cite{wei2022cot}, Tree-of-Thought (ToT)~\cite{yao2023tot}, and static and dynamic variants of a recent state-of-the-art adaptive reasoning framework, Meta-Reasoner \cite{sui2025meta}.

\paragraph{Models}
We evaluate across three different frontier models in order to test whether the \ourmethodshort\ controller transfers across backbone models without model-specific training: Gemini 3.1 Pro (high reasoning), Claude Sonnet 4.6 (high reasoning), and GPT-5.4 (medium reasoning).


\subsection{Main Results}

Table \ref{tab:main_results} reports accuracy across benchmarks, models, and reasoning methods. \ourmethodshort\ achieves the strongest average accuracy on all three models: 84.18\% on Gemini 3.1 Pro, 79.23\% on Claude Sonnet 4.6, and 71.84\% on GPT-5.4. The advantage is concentrated on the hardest benchmarks: relative to Direct, \ourmethodshort\ improves AIME by 43.33, 61.67, and 65.00 points on Gemini, Claude, and GPT-5.4 respectively, and improves LCB-Hard by 39.71, 44.29, and 9.14 points. These results suggest that adaptive inference is most useful when the task contains unresolved reasoning demands that cannot be addressed by a single forward pass.

The benefit of adaptive reasoning in \ourmethodshort\ scales sharply with task difficulty. The largest absolute gains occur on the hardest tasks: double-digit improvements on AIME (up to 65.00 points on GPT-5.4) and LCB-Hard confirm that adaptive computation is most valuable at the frontier of model capability. Gains on easier coding tasks are small by comparison-\ourmethodshort\ improves LCB-Easy by 3.42 points on Gemini 3.1 Pro and 0.93 points on Claude Sonnet 4.6, but falls marginally below Direct on GPT-5.4, which suggests little benefit, and slight overhead, when the model already solves the task directly. This is consistent with the design motivation of \ourmethodshort\: contest mathematics often requires identifying missing constraints, correcting partially useful solution paths, and allocating additional reasoning effort to unresolved bottlenecks.

Among the baselines, Tree-of-Thought (ToT) is most competitive on coding tasks, achieving the best LCB-Medium and LCB-Hard scores on both Claude Sonnet 4.6 and GPT-5.4 (where \ourmethodshort\ scores second-best, but scores below \ourmethod~on AIME (by 30.00 and 50.00 points) ) and on GPQA across all three models. One possible reading is that ToT's branching search suits problems with locally evaluable subgoals (e.g., passing test cases) more than tasks requiring global constraint satisfaction. Meta-Reasoner-Dynamic scores higher than its static counterpart on AIME (76.67\% $\xrightarrow{}$ 86.67\% on Claude Sonnet 4.6; 51.67\% $\xrightarrow{}$ 56.67\% on GPT-5.4) while remaining below \ourmethodshort, consistent with dynamic adaptation benefiting from the additional deficit-targeting signal \ourmethodshort~provides.

\begin{figure}[t]
  \centering
  \includegraphics[width=\textwidth]{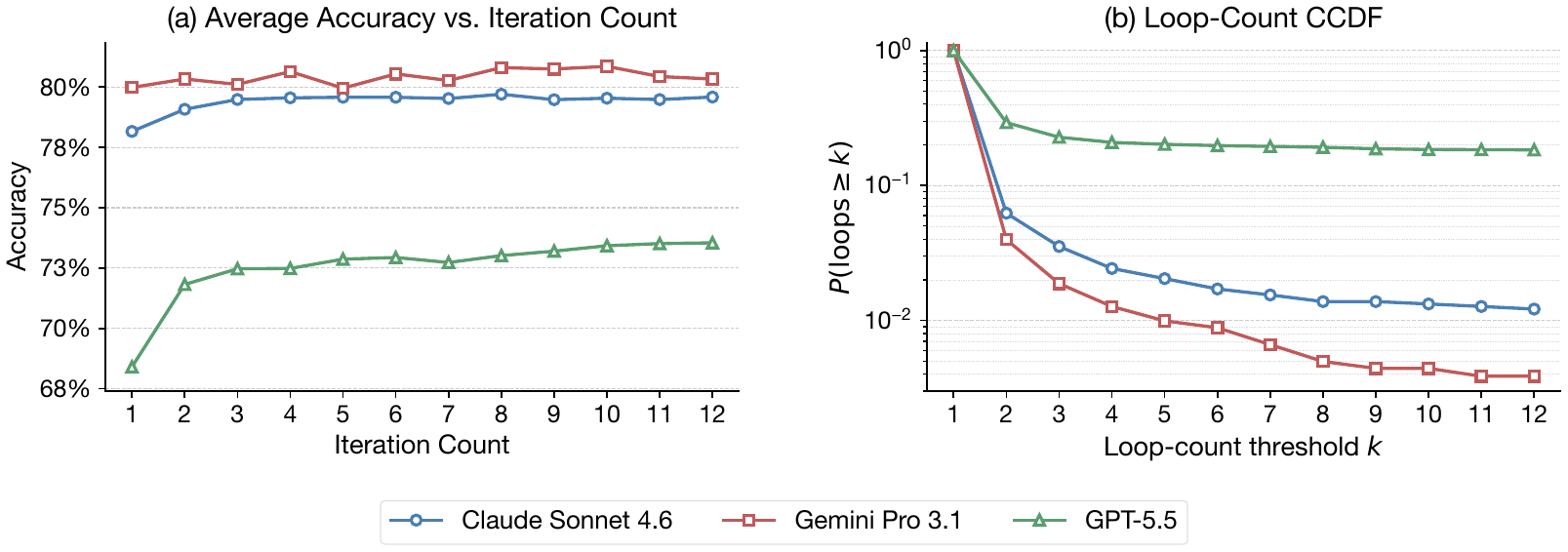}
  \caption{%
    \textbf{(a) Accuracy vs.\ iteration count} across three backbone models, measured by direct prompting with the context accumulated at each reasoning round. Accuracy rises with additional iterations before plateauing; GPT-5.5 benefits most, consistent with its heavier use of multiple rounds (panel~b).
    \textbf{(b) Complementary CDF of loop counts.}
    Gemini Pro 3.1 resolves the vast majority of example (${\approx}96\%$) in a single round, while GPT-5.5 exhibits a substantially heavier tail (${\approx}29\%$ require two or more rounds), explaining its larger accuracy gains with more iterations.
    Claude Sonnet 4.6 falls between the two extremes. }
  \label{fig:accuracy_and_distribution}
\end{figure}

\subsection{Iteration Scaling}

In order to test the effect of multiple rounds of reasoning, we consider the accuracy when generating final answers from traces truncated after each round. Reasoning traces from the evaluation were extracted from each round of the reasoning process and the model was prompted to produce a final answer based on the trace. As in the main evaluation, 12 total rounds of reasoning were allowed. 

Figure \ref{fig:accuracy_and_distribution} shows how the predictive power of the reasoning increases with the maximum number of iterations of the model. The iteration curves also show that gains can plateau. This suggests that additional reasoning is not always beneficial: once the key uncertainty has been resolved, further iterations may introduce redundant reasoning, overfitting to spurious concerns, or unnecessary revisions. This observation motivates the use of explicit progress evaluation and early stopping in \ourmethodshort.

GPT-5.4 and Claude Sonnet 4.6 show a clear increase in accuracy with the number of rounds of reasoning. The distribution of Gemini 3.1 Pro responses is dominated by fewer rounds of reasoning than the other models, as shown in Figure \ref{fig:accuracy_and_distribution} such that long reasoning processes do not contribute greatly to the mean accuracy. GPT-5.4 had more multi-round responses, and shows the greatest average benefit from multiple rounds of reasoning. Histograms showing the distribution of the number of rounds of reasoning can be found in Figure \ref{fig:accuracy_and_distribution}. The easy and medium difficulty splits of LiveCodeBench did not show a significant performance increase with the number of reasoning rounds, in contrast to the hard split that showed significant gains.

\section{Analysis}
\subsection{Ablation test: Demand ablation}

To quantify the contribution of demand tracking, we perform an ablation of demand tracking in \ourmethodshort. This ablation skips demand profiling, residual demand profiling, and does not inject demand information into context. It fixes the effort level to 5 (high) instead of calculating effort from the demand assessment. The ablated variant retains adaptive progress evaluation, effort scheduling, and iterative controller-worker interaction, all of which are part of the broader \ourmethodshort\ design.

On GPT-5.4, full \ourmethodshort\ averaged over the six datasets outperforms the ablated variant on each of the three backbone models. This is in spite of the fact that the ablated variant applies maximum reasoning effort at each step while \ourmethodshort\ dynamically adjusts effort based on task demands. This suggests that explicit demand tracking is useful to the meta-reasoning process.

\begin{table*}[t]
\centering
\caption{Ablation testing results. \ourmethodshort\ outperforms a demand-ablated maximum effort variant. The \ourmethodshort(ablated) method is identical to \ourmethodshort~except that it removes all demand profiling, residual demand assessment, and demand conditioned exemplars, and it fixes the effort level at 5 (maximum). Results are reported as percentage (\%) scores. LCB = LiveCodeBench (Easy/Medium/Hard). Best result per benchmark per model marked \textbf{bold}. }
\label{tab:ablation}
\resizebox{0.82\textwidth}{!}{%
\begin{tabular}{lcccccc|c}
\toprule
\textbf{Method} & \textbf{AIME} & \textbf{MATH500} & \textbf{GPQA} & \textbf{LCB-Easy} & \textbf{LCB-Med} & \textbf{LCB-Hard} & \textbf{Average} \\
\midrule
\multicolumn{8}{l}{\textit{Gemini 3.1 Pro}} \\
\midrule
\ourmethodshort\ (ablated)   & \textbf{100.00} & 90.80 & 93.94 & \textbf{82.92} & \textbf{65.27} & 70.57 & 83.92\\
\ourmethodshort\       & 98.33 & \textbf{91.40} & \textbf{95.45} & 82.30 &    65.01 &  \textbf{72.57} &  \textbf{84.18} \\
\midrule
\multicolumn{8}{l}{\textit{Claude Sonnet 4.6}} \\
\midrule
 \ourmethodshort\ (ablated)   & 93.33 & 87.40 & \textbf{82.83} & \textbf{84.16} & \textbf{64.23} & 60.29 & 78.71 \\
\ourmethodshort\       & \textbf{96.67} & \textbf{87.60} & 81.82 & \textbf{84.16} &    63.71 &  \textbf{61.43} & \textbf{79.23}\\
\midrule
\multicolumn{8}{l}{\textit{GPT-5.4}} \\
\midrule
 \ourmethodshort\ (ablated)   & 63.33 & \textbf{83.20} & 74.75 & \textbf{83.23} & \textbf{63.71} & 57.43 & 70.94\\
\ourmethodshort\       & \textbf{70.00} &    82.00 &    \textbf{75.25} &    81.68 &    62.66 & \textbf{59.43} & \textbf{71.84}\\
\bottomrule
\end{tabular}%
}
\end{table*}

\subsection{Example: Correcting a trajectory with adaptive reasoning}

We present an example of \ourmethodshort~from LiveCodeBench-Hard using a Gemini 3.1 Pro backbone to illustrate the ability of \ourmethodshort~to adapt reasoning and correct errors made earlier in the reasoning process.

\begin{figure}[t]
  \centering
  \includegraphics[width=0.85\textwidth]{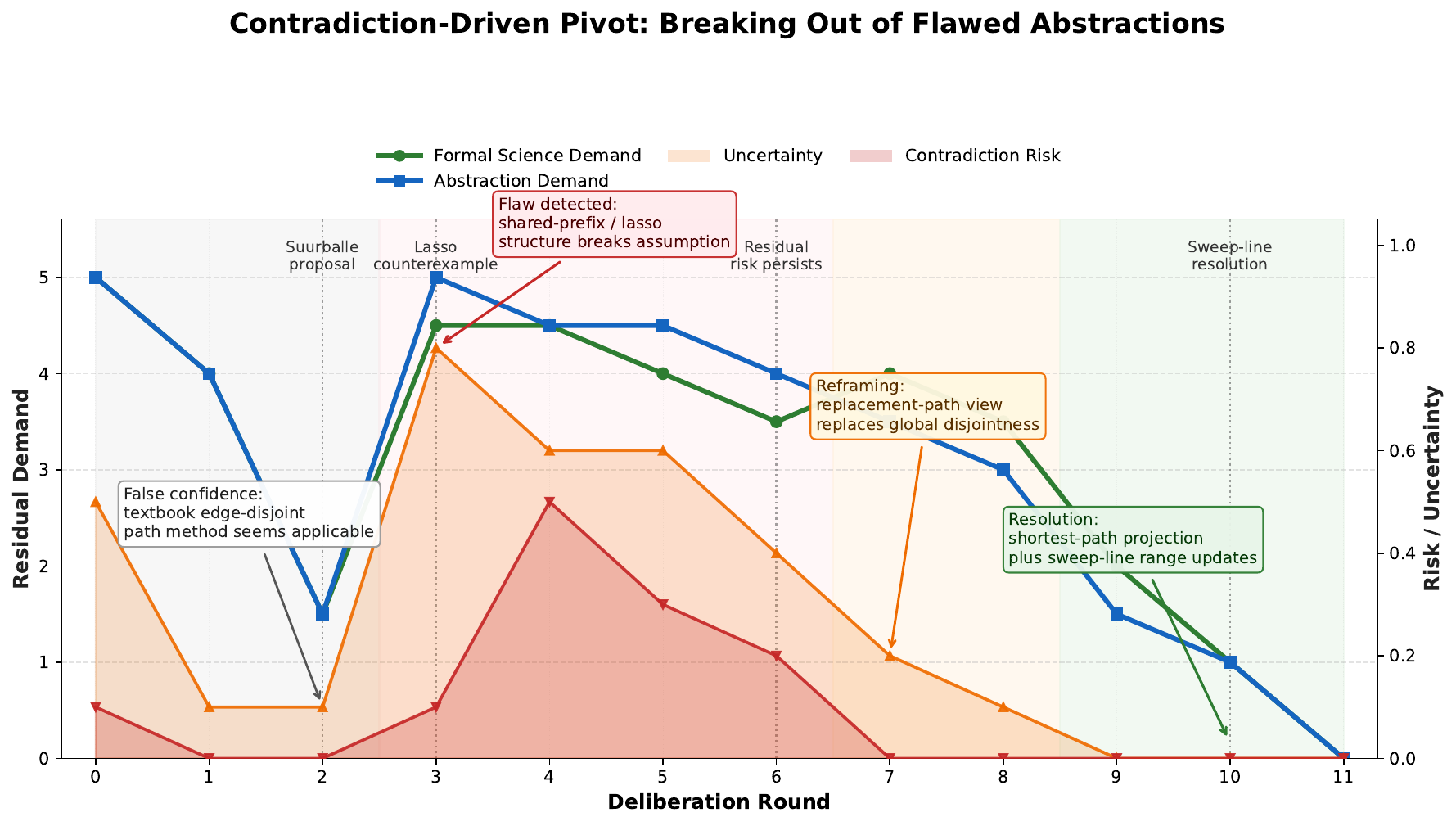}
  \caption{Demand-risk dynamics for arc191-d. The model initially attempts to solve the problem using familiar shortest-path machinery, leading to a temporary but misleading reduction in residual demand. Counterexamples involving shared paths and lasso-like graph structure expose the inadequacy of the Suurballe-style abstraction, producing a spike in uncertainty and contradiction risk. The eventual solution emerges from a replacement-path reframing with range updates over a shortest-path projection, after which both residual demand and uncertainty collapse.}
  \label{fig:arc191_d}
\end{figure}

To illustrate the Controller-Worker architecture's capacity for deep algorithmic critique, we analyze the \ourmethodshort\ trajectory on AtCoder \texttt{arc191\_d} (``Moving Pieces on Graph''). The problem requires finding the minimum moves to swap two pieces on a graph without them colliding, which geometrically requires finding a bypass structure (a vertex of degree $\ge 3$ or a cycle).

As shown in Figure~\ref{fig:arc191_d}, the \ourmethodshort\ reasoning undergoes a critical paradigm shift driven by its internal state tracking.

\textbf{Initial Textbook Application (Rounds 0--2):} 
\ourmethodshort\ correctly deduces that bypassing via a cycle requires finding two paths. It immediately maps this to a known formal algorithm, proposing: \textit{``We can find the minimum cost of two edge-disjoint paths using Suurballe's Algorithm.''} It confidently writes a full implementation using Johnson's potential transformation. Consequently, cognitive residuals and uncertainty drop to near zero.

\textbf{Critique and Contradiction Spike (Rounds 3--5):} 
During the progress assessment in Round 3, \ourmethodshort\ tests its own logic against structural edge cases and detects a fatal flaw. It notes: \textit{``Suurballe's algorithm strictly looks for two edge-disjoint paths from S to T. However, if the graph contains a bridge... the pieces can follow each other along the same path to reach a cycle, swap, and return.''} 
Recognizing that its model fails on ``lasso'' graphs, the \ourmethodshort\ uncertainty spikes to 0.8, and contradiction risk elevates, triggering an immediate halt to the generation process.

\textbf{Structural Discovery and Resolution (Rounds 6--11):} 
Forced to abandon Suurballe's algorithm, the Controller elevates the \textit{Abstraction} and \textit{Logical Reasoning} demands. \ourmethodshort\ pivots to analyzing the shortest path $P$ directly, deducing that the optimal cycle bypass cost is exactly $\min_{e \in P} (|P| + \text{dist}_{G \setminus e}(S, T))$. To avoid a Time Limit Exceeded (TLE) error, it further refines this into an $\mathcal{O}(M \log N)$ sweep-line algorithm using a Segment Tree with lazy deletions.

Traditional LLM decoding often falls victim to ``snowballing,'' where the model hallucinates during the reasoning process and downstream reasoning compounds the original error. By continuously calculating residual demand and contradiction risk, this architecture demonstrates the emergent ability to invalidate its own assumptions, discard heavily invested reasoning paths, and discover the correct solution to problems.

%% file: sections/conclusion.tex
\section{Conclusion}

\par
We presented Cognitive Demand Steering (CDS), a training-free technique for adaptive reasoning at inference time. CDS profiles the task along a set of cognitively grounded dimensions and tracks the remaining demand along these dimensions over multiple iterations, using this profile to choose controller actions and plan effort. CDS maintains a representation of what remains unresolved in the task and steers the reasoning trajectory accordingly.

Across mathematics, scientific question answering, and coding benchmarks, CDS achieves the best average accuracy across the evaluated methods, though gains vary by benchmark and are smaller or occasionally negative on easier coding tasks. Its strongest gains appear on difficult mathematics and hard coding tasks, while on easier coding tasks the benefits are smaller and sometimes outweighed by overhead. Iteration-scaling results show that additional reasoning rounds are most valuable when the task is complex or demanding, supporting the motivation for demand-aware adaptive inference. Ablation of demand tracking suggests that the residual demand tracking mechanism is useful but perhaps not yet uniformly beneficial, motivating future work on demand tracking and its linkage to the controller.

CDS provides an easily interpretable framework for improved inference time reasoning, pointing towards inference systems that, instead of simply reasoning more, target reasoning towards specific unresolved demands. Instead of simply asking "What action should be taken next?" CDS asks "What action will resolve the remaining demands of this task?"

\section*{Limitations}

\par
Although CDS can produce substantially improved performance, there are some limitations that are worth mentioning. First, CDS introduces significant computational overhead during inference. The various components (progress assessor, controller, etc.) all require separate LLM calls, resulting in additional latency and a higher token budget. This additional cost could be less impactful when using CDS to train an LLM rather than when performing pure inference.

\par
The performance of CDS depends on the performance of the underlying LLM to a significant extent. If the LLM can't make accurate assessments of progress or select meaningful actions given a relatively long context, then the internal mechanisms of CDS will not be able to guide the reasoning process effectively. 

\par
The demand ablation suggests that simply exposing the demand profile to the CDS controller may not always be sufficient. In some cases it may expand the context window without meaningfully improving performance. In such a case, or if the token budget is restrictive, the ablated version of CDS without demand tracking may produce similar performance with fewer tokens.

\par
The demand profiler relies on a fixed taxonomy and hand-authored exemplars, both grounded in assumptions about task structure that may not generalize. For out-of-distribution tasks - such as highly specialized domains or highly multimodal inputs - calibration of general scales (\cite{zhou2025generalscales} on highly specialized or multimodel inputs has not yet been assessed. We have evaluated CDS with textual inputs. More evaluation is needed for multimodal inputs, tool use, and interactive environments, and the set of cognitive demands might need to be extended to accommodate new modalities.

%% file: sections/appendix.tex
\appendix

\section{Technical appendices and supplementary material}

\subsection{Prompts}

The prompts used in the CDS pipeline can be found here.

\begin{llmlogbox}{Demand Profiler Prompt}
Score the following task on each of the 16 cognitive / knowledge dimensions with an integer from 0 to 5.
Base the scores only on the current example.

Dataset: {dataset}
Subject: {subject or 'n/a'}
Task instance:
{task_text}

Dimensions:
    - attention_scan: Score 0-5 for how much the task requires locating target information among distractors, scanning text/context, or tracking relevant items.
    - calibrating_knowns_unknowns: Score 0-5 for how much the task requires explicitly separating what is given, inferred, uncertain, or missing before solving.
    - abstraction: Score 0-5 for how much the task requires forming abstractions, schemas, variables, symbolic representations, or general principles.
    - critical_thinking: Score 0-5 for how much the task requires checking assumptions, rejecting distractors, validating evidence, or questioning a tempting answer.
    - identify_relevant_information: Score 0-5 for how much the task requires selecting which facts, sentences, constraints, or evidence matter for the answer.
    - knowledge_applied_science: Score 0-5 for how much the task needs applied scientific or technical knowledge such as programming, engineering, or software debugging.
    - knowledge_customary: Score 0-5 for how much the task depends on everyday/common world knowledge rather than specialized study.
    - knowledge_formal_science: Score 0-5 for how much the task depends on formal math, logic, statistics, symbolic manipulation, or algorithmic knowledge.
    - knowledge_natural_science: Score 0-5 for how much the task depends on biology, chemistry, physics, astronomy, or other natural sciences.
    - knowledge_social_science: Score 0-5 for how much the task depends on economics, law, politics, history, sociology, psychology, or similar domains.
    - logical_reasoning: Score 0-5 for how much the task requires chaining implications, proving consistency, case analysis, or multi-step deduction.
    - mind_modelling: Score 0-5 for how much the task requires reasoning about beliefs, intentions, social roles, incentives, or perspective taking.
    - quantitative_reasoning: Score 0-5 for how much the task requires arithmetic, algebra, estimation, numerical comparison, probability, or quantitative derivation.
    - spatial_reasoning: Score 0-5 for how much the task requires geometric, topological, coordinate-based, or other spatial reasoning.
    - verbal_comprehension: Score 0-5 for how much the task requires careful reading, paraphrase, reference resolution, text integration, or nuanced interpretation.
    - verbal_expression: Score 0-5 for how much successful completion depends on clearly expressing the reasoning or answer in a specific linguistic format.

Return strict JSON with keys:
{{
  "scores": {{"dimension": 0}},
  "top_dimensions": ["dim1", "dim2", "dim3", "dim4"],
  "top_domains": ["knowledge_x", "knowledge_y"],
  "effort_tier": 1,
  "notes": "one short sentence"
}}

\end{llmlogbox}

\begin{llmlogbox}{Residual Demand Assessment prompt}
Assess the remaining cognitive demand on the task after the current worker reasoning.

Task:
{task_text}

Current reasoning / draft answer:
{reasoning_so_far or '[empty]'}

Candidate dimensions:
    Candidate dimensions:
    - attention_scan: initial_demand=<value>
    - calibrating_knowns_unknowns: initial_demand=<value>
    - abstraction: initial_demand=<value>
    - critical_thinking: initial_demand=<value>
    - identify_relevant_information: initial_demand=<value>
    - knowledge_applied_science: initial_demand=<value>
    - knowledge_customary: initial_demand=<value>
    - knowledge_formal_science: initial_demand=<value>
    - knowledge_natural_science: initial_demand=<value>
    - knowledge_social_science: initial_demand=<value>
    - logical_reasoning: initial_demand=<value>
    - mind_modelling: initial_demand=<value>
    - quantitative_reasoning: initial_demand=<value>
    - spatial_reasoning: initial_demand=<value>
    - verbal_comprehension: initial_demand=<value>
    - verbal_expression: initial_demand=<value>

Return strict JSON with keys:
{{
  "residuals": {{"dimension": 0.0}},
  "notes": "one short sentence"
}}

Residual values must be floats in [0,5], where:
- 0 means essentially no meaningful demand remains on that dimension
- 5 means very high remaining demand on that dimension

Only return residuals for the listed candidate dimensions.
Do not omit any listed dimension.

\end{llmlogbox}

\begin{llmlogbox}{Progress Evaluator prompt}
Evaluate the current reasoning progress on the task.

Task:
{task_text}

Current reasoning / draft answer:
{reasoning_so_far or '[empty]'}

Active dimensions with residual demand and derived coverage:
{dim_block}

Return strict JSON with keys:
{{
  "solved": false,
  "answer_candidate": null,
  "uncertainty": 0.0,
  "loop_risk": 0.0,
  "contradiction_risk": 0.0,
  "missing_information": ["Concrete missing fact, value, constraint, or intermediate result listed here"],
  "verification_targets": ["Verification targets listed here"],
  "critique": "one concise paragraph"
}}

Do not return a coverage field. Coverage has already been computed from residual demand.
missing_information must be a list. Use [] only if no missing facts, values, constraints, or intermediate results remain.
verification_targets must be a list. Use [] only if no verification checks remain.

\end{llmlogbox}

\begin{llmlogbox}{Controller prompt}
You are the meta-reasoner. Choose the next concrete action for the worker.

Task:
{task_text}{demand_section}
Progress state:
{json.dumps(progress_json, ensure_ascii=False)}

Missing information from progress assessment:
{json.dumps(missing_information, ensure_ascii=False)}

Existing verification targets from progress assessment:
{json.dumps(verification_targets, ensure_ascii=False)}

Effort tier: {effort_tier} ({EFFORT_TIERS[effort_tier]['label']})
Dataset hint: {hint}

Dimension guidance exemplars:
{exemplar_block}

Return strict JSON as a single object, not null:
{{
  "reflection": "short description of what is working or stalled",
  "profile_alignment": "how the next step should address the active residual dimensions",
  "missing_information": ["specific missing fact, value, constraint, or intermediate result still needed"],
  "effort_adjustment": "increase|keep|decrease",
  "next_action": "exactly one concrete action sentence for the worker",
  "verification_targets": ["specific check to perform", "another specific check if needed"],
  "style_constraints": ["constraint for the worker"]
}}

Rules:
- Do not return null.
- Do not omit next_action.
- missing_information must be a list. Use [] only if no missing facts, values, constraints, or intermediate results remain.
- verification_targets must be a list. Use [] only if no verification checks remain.
- If the progress assessment already listed missing_information, preserve or refine it unless it is clearly resolved.
- If the next action depends on deriving, extracting, estimating, or checking something, include that item in missing_information or verification_targets, as appropriate.    

\end{llmlogbox}

\begin{llmlogbox}{Worker prompt (for coding tasks)}
Task:
{task_text}

Existing reasoning:
{reasoning_so_far or '[empty]'}

Meta-guidance:
{json.dumps(controller_json, ensure_ascii=False)}

Effort budget:
- tier: {effort_tier}
- max_new_steps: {settings['max_new_steps']}
- branch_budget: {settings['branch_budget']}
- verification_mode: {settings['verification']}

Write only the next chunk of reasoning, not a fresh restart unless explicitly asked.
Be concise and action-oriented. If you are ready to propose a solution, end with:
FINAL ANSWER:
```python
# your solution
```
The fenced block must contain only runnable Python 3 code. If you are not completely ready to propose a solution, only output your reasoning about the current step.

\end{llmlogbox}

\begin{llmlogbox}{Worker prompt (for non-coding tasks)}
Task:
{task_text}

Existing reasoning:
{reasoning_so_far or '[empty]'}

Meta-guidance:
{json.dumps(controller_json, ensure_ascii=False)}

Effort budget:
- tier: {effort_tier}
- max_new_steps: {settings['max_new_steps']}
- branch_budget: {settings['branch_budget']}
- verification_mode: {settings['verification']}

Write only the next chunk of reasoning, not a fresh restart unless explicitly asked.
Be concise and action-oriented. If you are not completely ready to propose a solution, only output your reasoning about the current step.

\end{llmlogbox}

\subsection{Cognitive scale dimensions.}
\label{app:dimensions}
We characterize each evaluation item using the cognitive scale dimensions adopted in
Scales++ \citep{bean2025scalespp}. These dimensions operationalize the
cognitive and knowledge-based demands of a task instance, providing an interpretable
item-level representation of what abilities are required for successful completion.
Each task instance can be rated on a 0--5 scale for each dimension, where higher
values indicate a greater contribution of that ability to solving the task.

\begin{table}[htbp]
\centering
\caption{Cognitive scale dimensions used for task-level annotation, adapted from Appendix B of Scales++.}
\label{tab:cognitive-scale-dimensions}
\begin{tabular}{ll}
\toprule
\textbf{Abbreviation} & \textbf{Dimension} \\
\midrule
AS   & Attention and scan \\
CKU  & Calibrating knowns and unknowns \\
CLA  & Conceptualisation, learning, and abstraction \\
CTP  & Critical thinking processes \\
IRI  & Identifying relevant information \\
KAS  & Knowledge: applied science \\
KC   & Knowledge: customary \\
KFS  & Knowledge: formal science \\
KNS  & Knowledge: natural science \\
KSS  & Knowledge: social science \\
LR   & Logical reasoning \\
MMSC & Mind modelling and social cognition \\
QR   & Quantitative reasoning \\
SRN  & Spatial reasoning and navigation \\
VC   & Verbal comprehension \\
VE   & Verbal expression \\
\bottomrule
\end{tabular}
\end{table}

\subsection{Exemplars}
\label{subsec:exemplars}

General-purpose exemplars are included as in-context guidance when particular dimensions are active. Each exemplar consists of a ``hint'' providing advice on a particular cognitive dimension as well as a ``pitfall'' warning regarding common failure modes.

\begin{llmlogbox}{Exemplar: Attention and scan}
hint:
Slow down and locate the exact target information before reasoning further.
Scan the task for constraints, definitions, quantities, edge cases, and distractors,
separate what is explicitly stated from what is merely nearby or tempting.

pitfall:
Reasoning from a salient but non-target detail, overlooking a condition, or losing track 
of information spread across the prompt.

\end{llmlogbox}

\begin{llmlogbox}{Exemplar: Verbal comprehension}
hint:
Restate the task in simpler terms and resolve references, qualifiers, and hidden conditions. 
Identify exactly what is being asked, what each phrase rules in or out, and whether any wording 
changes the mathematical, scientific, or coding interpretation.

pitfall:
Solving a nearby problem because a key phrase, quantifier, exception, or referent was misread.

\end{llmlogbox}

\begin{llmlogbox}{Exemplar: Verbal expression}
hint:
Format the final response so it directly matches the requested output. Use concise, unambiguous 
language; avoid burying the answer in a derivation; and make the final answer easy to parse.

pitfall:
Having the right solution but expressing it in a vague, overlong, malformed, or unparseable way.

\end{llmlogbox}

\begin{llmlogbox}{Exemplar: Abstraction}
hint:
Identify the underlying structure of the problem. Replace surface details with variables, 
invariants, cases, recurrence structure, state transitions, or a general schema before computing.

pitfall:
Staying too close to surface wording and missing the reusable pattern or simplifying abstraction.

\end{llmlogbox}

\begin{llmlogbox}{Exemplar: Identify relevant information}
hint:
List the facts that are necessary for the next inference and explicitly discard irrelevant or 
distracting details. Check that every used fact supports the target quantity, answer choice, or code
behavior.

pitfall:
Using information because it is present rather than because it is relevant to the requested answer.

\end{llmlogbox}

\begin{llmlogbox}{Exemplar: Critical thinking}
hint:
Audit the reasoning before advancing. Look for unsupported assumptions, fragile steps, off-by-one
errors, unit mismatches, invalid simplifications, and whether an alternative interpretation could break
the answer.

pitfall:
Continuing a plausible path without checking whether its assumptions or intermediate results are valid.

\end{llmlogbox}

\begin{llmlogbox}{Exemplar: Calibrating knowns and unknowns}
hint:
Separate known facts, inferred facts, assumptions, and unknowns. Mark any uncertain step explicitly,
then decide whether to verify, derive, estimate, or avoid relying on it.

pitfall:
Treating a guess or unstated assumption as established fact.

\end{llmlogbox}

\begin{llmlogbox}{Exemplar: Mind modelling }
hint:
Model the relevant agents, intentions, beliefs, incentives, or perspectives. Track who knows what,
what each actor is trying to achieve, and how social or strategic context changes the conclusion.

pitfall:
Answering from the solver's perspective when the task depends on another agent's beliefs or goals.

\end{llmlogbox}

\begin{llmlogbox}{Exemplar: Logical reasoning}
hint:
Make the inference chain explicit. State premises, derive consequences step by step, check cases,
and verify that the conclusion follows from the constraints rather than from pattern matching.

pitfall:
Skipping a necessary implication, merging cases incorrectly, or assuming the desired conclusion.

\end{llmlogbox}

\begin{llmlogbox}{Exemplar: Quantitative reasoning}
hint:
Define the quantities, equations, units, and target value before calculating. Preserve exact forms 
when needed, check arithmetic, and compare the magnitude of the result against the problem context.

pitfall:
Computing the wrong quantity, losing units, rounding too early, or accepting an implausible magnitude.

\end{llmlogbox}

\begin{llmlogbox}{Exemplar: Spatial reasoning }
hint:
Construct an explicit spatial representation: diagram mentally, name coordinates or regions, track 
relative positions, and reason about transformations, containment, distance, adjacency, or geometry.

pitfall:
Relying on vague visual intuition without preserving the exact spatial relationships.

\end{llmlogbox}

\begin{llmlogbox}{Exemplar: Knowledge - Applied science }
hint:
Identify the applied domain principle involved, such as engineering, medicine, computing, law, or
professional practice. 
Use the domain rule carefully, then connect it to the specific facts of the task.

pitfall:
Using everyday intuition where a technical or professional rule is needed.

\end{llmlogbox}

\begin{llmlogbox}{Exemplar: Knowledge - Formal science }
hint:
Recall the relevant formal tool: theorem, identity, algorithm, proof pattern, statistical rule, or
symbolic transformation. 
Apply it exactly and verify that its preconditions hold.

pitfall:
Applying a familiar formula or theorem without checking that the situation satisfies its assumptions.

\end{llmlogbox}

\begin{llmlogbox}{Exemplar: Knowledge - Natural science }
hint:
Identify the governing natural-science concept, mechanism, law, or empirical relation. Keep track of 
scale, 
units, causal direction, and whether the problem asks for explanation, prediction, or calculation.

pitfall:
Mixing up similar scientific concepts or using a qualitative fact where a quantitative relation is 
required.

\end{llmlogbox}

\begin{llmlogbox}{Exemplar: Knowledge - Social science }
hint:
Identify the relevant social-science or humanities framework: historical context, legal rule, economic
incentive, 
psychological mechanism, philosophical distinction, or cultural interpretation. Apply it to the
specific case.

pitfall:
Treating a socially situated or historically specific question as if it were context-free.

\end{llmlogbox}

\begin{llmlogbox}{Customary knowledge }
hint:
Use ordinary world knowledge cautiously and only where it is relevant. Make explicit which commonsense 
facts are being assumed, and check whether the task context overrides ordinary expectations.

pitfall:
Overgeneralizing from commonsense expectations when the prompt gives a more specific condition.

\end{llmlogbox}

%% file: neurips_2026.bbl
\begin{thebibliography}{10}

\bibitem{bean2025scalespp}
Andrew~M. Bean, Nabeel Seedat, Shengzhuang Chen, and Jonathan~Richard Schwarz.
\newblock Scales++: Compute efficient evaluation subset selection with cognitive scales embeddings.
\newblock {\em arXiv preprint arXiv:2510.26384}, 2025.

\bibitem{chen2024not}
Xingyu Chen, Jiahao Xu, Tian Liang, Zhiwei He, Jianhui Pang, Dian Yu, Linfeng Song, Qiuzhi Liu, Mengfei Zhou, Zhuosheng Zhang, et~al.
\newblock Do not think that much for 2+ 3=? on the overthinking of o1-like llms.
\newblock {\em arXiv preprint arXiv:2412.21187}, 2024.

\bibitem{cuesta2025large}
Jhouben Cuesta-Ramirez, Samuel Beaussant, and Mehdi Mounsif.
\newblock Large reasoning models are not thinking straight: on the unreliability of thinking trajectories.
\newblock {\em arXiv preprint arXiv:2507.00711}, 2025.

\bibitem{deepseek2025r1}
{DeepSeek-AI}.
\newblock Deepseek-r1: Incentivizing reasoning capability in {LLMs} via reinforcement learning.
\newblock {\em Nature}, 645:633--638, 2025.

\bibitem{gao2024meta}
Peizhong Gao, Ao~Xie, Shaoguang Mao, Wenshan Wu, Yan Xia, Haipeng Mi, and Furu Wei.
\newblock Meta reasoning for large language models.
\newblock {\em arXiv preprint arXiv:2406.11698}, 2024.

\bibitem{livecodebench}
Naman Jain, King Han, Alex Gu, Wen-Ding Li, Fanjia Yan, Tianjun Zhang, Sida Wang, Armando Solar-Lezama, Koushik Sen, and Ion Stoica.
\newblock Livecodebench: Holistic and contamination free evaluation of large language models for code.
\newblock {\em arXiv preprint arXiv:2403.07974}, 2024.

\bibitem{jiang2026chain}
Tianyi Jiang, Arctanx An, Hengyi Feng, Naixin Zhai, Haodong Li, Xiaomin Yu, Jiahui Liu, Hanwen Du, Shuo Zhang, Zhi Yang, Jie Huang, Youhua Li, Yongxin Ni, Huacan Wang, and Ronghao Chen.
\newblock Chain of mindset: Reasoning with adaptive cognitive modes.
\newblock {\em arXiv preprint arXiv:2602.10063}, 2026.

\bibitem{kargupta2025cognitive}
Priyanka Kargupta, Shuyue~Stella Li, Haocheng Wang, Jinu Lee, Shan Chen, Orevaoghene Ahia, Dean Light, Thomas~L Griffiths, Max Kleiman-Weiner, Jiawei Han, et~al.
\newblock Cognitive foundations for reasoning and their manifestation in llms.
\newblock {\em arXiv preprint arXiv:2511.16660}, 2025.

\bibitem{lightman2024let}
Hunter Lightman, Vineet Kosaraju, Yuri Burda, Harrison Edwards, Bowen Baker, Teddy Lee, Jan Leike, John Schulman, Ilya Sutskever, and Karl Cobbe.
\newblock Let's verify step by step.
\newblock In {\em International Conference on Learning Representations}, volume 2024, pages 39578--39601, 2024.

\bibitem{liu2025metascale}
Qin Liu, Wenxuan Zhou, Nan Xu, James~Y Huang, Fei Wang, Sheng Zhang, Hoifung Poon, and Muhao Chen.
\newblock Metascale: Test-time scaling with evolving meta-thoughts.
\newblock {\em arXiv preprint arXiv:2503.13447}, 2025.

\bibitem{madaan2023selfrefine}
Aman Madaan, Niket Tandon, Prakhar Gupta, Skyler Hallinan, Luyu Gao, Sarah Wiegreffe, Uri Alon, Nouha Dziri, Shrimai Prabhumoye, Yiming Yang, Shashank Gupta, Bodhisattwa~Prasad Majumder, Katherine Hermann, Sean Welleck, Amir Yazdanbakhsh, and Peter Clark.
\newblock Self-refine: Iterative refinement with self-feedback.
\newblock {\em arXiv preprint arXiv:2303.17651}, 2023.

\bibitem{openai2024reasoning}
{OpenAI}.
\newblock Learning to reason with {LLMs}.
\newblock \url{https://openai.com/index/learning-to-reason-with-llms/}, 2024.
\newblock Accessed: 2026-07-29.

\bibitem{gpqa}
David Rein, Betty~Li Hou, Asa~Cooper Stickland, Jackson Petty, Richard~Yuanzhe Pang, Julien Dirani, Julian Michael, and Samuel~R Bowman.
\newblock Gpqa: A graduate-level google-proof q\&a benchmark.
\newblock {\em arXiv preprint arXiv:2311.12022}, 2023.

\bibitem{shinn2023reflexion}
Noah Shinn, Federico Cassano, Edward Berman, Ashwin Gopinath, Karthik Narasimhan, and Shunyu Yao.
\newblock Reflexion: Language agents with verbal reinforcement learning.
\newblock {\em arXiv preprint arXiv:2303.11366}, 2023.

\bibitem{snell2025scaling}
Charlie~Victor Snell, Jaehoon Lee, Kelvin Xu, and Aviral Kumar.
\newblock Scaling llm test-time compute optimally can be more effective than scaling parameters for reasoning.
\newblock In {\em The Thirteenth International Conference on Learning Representations}, 2025.

\bibitem{song2026large}
Peiyang Song, Pengrui Han, and Noah Goodman.
\newblock Large language model reasoning failures.
\newblock {\em arXiv preprint arXiv:2602.06176}, 2026.

\bibitem{su2025between}
Jinyan Su, Jennifer Healey, Preslav Nakov, and Claire Cardie.
\newblock Between underthinking and overthinking: An empirical study of reasoning length and correctness in llms.
\newblock {\em arXiv preprint arXiv:2505.00127}, 2025.

\bibitem{sui2025meta}
Yuan Sui, Yufei He, Tri Cao, Simeng Han, Yulin Chen, and Bryan Hooi.
\newblock Meta-reasoner: Dynamic guidance for optimized inference-time reasoning in large language models.
\newblock {\em arXiv preprint arXiv:2502.19918}, 2025.

\bibitem{wang2022selfconsistency}
Xuezhi Wang, Jason Wei, Dale Schuurmans, Quoc Le, Ed~Chi, Sharan Narang, Aakanksha Chowdhery, and Denny Zhou.
\newblock Self-consistency improves chain of thought reasoning in language models.
\newblock {\em arXiv preprint arXiv:2203.11171}, 2022.

\bibitem{wei2022cot}
Jason Wei, Xuezhi Wang, Dale Schuurmans, Maarten Bosma, Brian Ichter, Fei Xia, Ed~Chi, Quoc Le, and Denny Zhou.
\newblock Chain-of-thought prompting elicits reasoning in large language models.
\newblock {\em arXiv preprint arXiv:2201.11903}, 2022.

\bibitem{wu2025efficiency}
Chao Wu, Baoheng Li, Mingchen Gao, Yu~Tian, and Zhenyi Wang.
\newblock From efficiency to adaptivity: A deeper look at adaptive reasoning in large language models.
\newblock {\em arXiv preprint arXiv:2511.10788}, 2025.

\bibitem{yan2025position}
Hanqi Yan, Linhai Zhang, Jiazheng Li, Zhenyi Shen, and Yulan He.
\newblock Position: Llms need a bayesian meta-reasoning framework for more robust and generalizable reasoning.
\newblock In {\em Forty-second International Conference on Machine Learning Position Paper Track}, 2025.

\bibitem{yang2406buffer}
Ling Yang, Zhaochen Yu, Tianjun Zhang, Shiyi Cao, Minkai Xu, Wentao Zhang, Joseph~E Gonzalez, and Bin Cui.
\newblock Buffer of thoughts: thought-augmented reasoning with large language models (2024).
\newblock {\em URL https://arxiv. org/abs/2406.04271}, 2024.

\bibitem{yao2023tot}
Shunyu Yao, Dian Yu, Jeffrey Zhao, Izhak Shafran, Thomas~L. Griffiths, Yuan Cao, and Karthik Narasimhan.
\newblock Tree of thoughts: Deliberate problem solving with large language models.
\newblock {\em arXiv preprint arXiv:2305.10601}, 2023.

\bibitem{zhang2025survey}
Qiyuan Zhang, Fuyuan Lyu, Zexu Sun, Lei Wang, Weixu Zhang, Wenyue Hua, Haolun Wu, Zhihan Guo, Yufei Wang, Niklas Muennighoff, et~al.
\newblock A survey on test-time scaling in large language models: What, how, where, and how well?
\newblock {\em arXiv preprint arXiv:2503.24235}, 2025.

\bibitem{zhou2022leasttomost}
Denny Zhou, Nathanael Sch{\"a}rli, Le~Hou, Jason Wei, Nathan Scales, Xuezhi Wang, Dale Schuurmans, Claire Cui, Olivier Bousquet, Quoc Le, and Ed~Chi.
\newblock Least-to-most prompting enables complex reasoning in large language models.
\newblock {\em arXiv preprint arXiv:2205.10625}, 2022.

\bibitem{zhou2025generalscales}
Lexin Zhou, Lorenzo Pacchiardi, Fernando Mart{\'i}nez-Plumed, Katherine~M. Collins, Yael Moros-Daval, Seraphina Zhang, Qinlin Zhao, Yitian Huang, Luning Sun, Jonathan~E. Prunty, Zongqian Li, Pablo S{\'a}nchez-Garc{\'i}a, Kexin~Jiang Chen, Pablo A.~M. Casares, Jiyun Zu, John Burden, Behzad Mehrbakhsh, David Stillwell, Manuel Cebrian, Jindong Wang, Peter Henderson, Sherry~Tongshuang Wu, Patrick~C. Kyllonen, Lucy Cheke, Xing Xie, and Jos{\'e} Hern{\'a}ndez-Orallo.
\newblock General scales unlock ai evaluation with explanatory and predictive power.
\newblock {\em arXiv preprint arXiv:2503.06378}, 2025.

\bibitem{zhu2025scaling}
King Zhu, Hanhao Li, Siwei Wu, Tianshun Xing, Dehua Ma, Xiangru Tang, Minghao Liu, Jian Yang, Jiaheng Liu, Yuchen~Eleanor Jiang, et~al.
\newblock Scaling test-time compute for llm agents.
\newblock {\em arXiv preprint arXiv:2506.12928}, 2025.

\bibitem{zhuang2026beyond}
Ziqing Zhuang, Linhai Zhang, Jiasheng Si, Deyu Zhou, and Yulan He.
\newblock Beyond meta-reasoning: Metacognitive consolidation for self-improving llm reasoning.
\newblock {\em arXiv preprint arXiv:2604.17399}, 2026.

\end{thebibliography}
